\documentclass{article}

\usepackage[preprint, nonatbib]{neurips_2023}

\usepackage[utf8]{inputenc} %
\usepackage[T1]{fontenc}    %
\usepackage{hyperref}       %
\usepackage{url}            %
\usepackage{booktabs}       %
\usepackage{amsfonts}       %
\usepackage{nicefrac}       %
\usepackage{microtype}      %
\usepackage{xcolor}         %
\usepackage[pdftex]{graphicx}

\title{Behavioral Cloning via Search in Embedded Demonstration Dataset}

\begin{document}

\author{
Federico Malato\footnote{Equal contribution.} \\
School of Computing \\
University of Eastern Finland\\
Joensuu, Finland \\
\texttt{federico.malato@uef.fi}
\And
Florian Leopold$^{*}$ \\
Bielefeld University \\
Bielefeld, Germany \\
\texttt{fleopold@techfak.uni-bielefeld.de}
\And
Ville Hautam\"aki \\
School of Computing \\
University of Eastern Finland \\
Joensuu, Finland \\
\texttt{villeh@cs.uef.fi}
\And
Andrew Melnik \\
Bielefeld University \\
Bielefeld, Germany \\
\texttt{andrew.melnik.papers@gmail.com}
}
\maketitle

\begin{abstract}
Behavioural cloning uses a dataset of demonstrations to learn a behavioural policy. To overcome various learning and policy adaptation problems, we propose to use latent space to index a demonstration dataset, instantly access similar relevant experiences, and copy behavior from these situations. Actions from a selected similar situation can be performed by the agent until representations of the agent's current situation and the selected experience diverge in the latent space. Thus, we formulate our control problem as a search problem over a dataset of experts' demonstrations. We test our approach on BASALT MineRL-dataset in the latent representation of a Video PreTraining model. We compare our model to state-of-the-art Minecraft agents. Our approach can effectively recover meaningful demonstrations and show human-like behavior of an agent in the Minecraft environment in a wide variety of scenarios. Experimental results reveal that performance of our search-based approach is comparable to trained models, while allowing zero-shot task adaptation by changing the demonstration examples.

\end{abstract}

\section{Introduction}

\textit{Imitation Learning} (IL)~\cite{Schaal1996} and \textit{Reinforcement Learning} (RL)~\cite{SuttonBarto1998,schilling2019approach,bach2020learn,schilling2021decentralized} are commonly used methods to train agents to perform tasks in simulated and real environments, including multi-skill agents. Despite the extensive research in these fields, there are persistent challenges that pose difficulties for IL and RL methods. Firstly, these methods typically require significant computational resources for training. Secondly, when faced with new experiences, retraining or fine-tuning of trained models becomes necessary. Thirdly, the lack of one-shot or few-shot adaptability is a bottleneck, which is current focus of numerous studies on large language \cite{ouyang2022training} and vision models \cite{oquab2023dinov2, rana2023contrastive}. Exploring alternative approaches to IL and RL methods for control problems can potentially address these challenges and yield advantages in specific application domains.

A multi-modal latent space can be used for measuring similarity and multi-modal search. Some common ways to achieve multi-modal mapping is \textit{contrastive learning} \cite{radford2021learning} and \textit{supervised learning}, where in the later case representation at the last layers or a few layers before the output can serve as a multi-modal latent space \cite{baker2022video}. For control tasks, such multi-modal latent space can be used to connect visual and action experiences \cite{beohar2022solving} based on a demonstration dataset. Thus, database search in this latent space can provide actions for control. Such an approach can lead to improvements with respect to the three general challenges for IL and RL approaches mentioned above.

\paragraph{Motivation}
The present study is motivated by the MineRL BASALT 2022 challenge \cite{ShahDragan2021, Milani2023}. In the challenge, an agent must solve the four tasks: find a cave, build an animal pen, build a village house, and make a waterfall~\cite{ShahDragan2021, Milani2023}. For these tasks, no reward function is provided. Agents were judged by human contractors on two criteria: success in the task and human-likeliness of behaviour. Although, the objective is not mathematically well-defined, for a human, the success in completing such tasks is easy to evaluate and select the best one out of two performance examples. To solve the tasks, participants had a dataset of expert demonstrations at their disposal. Each demonstration trajectory of image-action pairs contains a single episode of one of the tasks solved by a human expert. Simplest solution to this then is {\em behavioral cloning} (BC), where policy is trained using supervised learning. 

BC is limited by the expert demonstration trajectories used in the training, and due to the training scheme used, sampled action per state is optimal in terms of expectation. But in practice agents are facing different {\em situations} throughout the episode, where one might be avoiding an obstacle and other to search for suitable location. These situations each then clearly require different solutions. 

To this end, we introduce {\em search-based behavioural cloning} (S-BC), an approach to BC that addresses that issue. S-BC generates a latent space of experts' trajectories that encapsulate both present and past information, and reformulates the control problem as a search problem in such a space.  By allowing to search for {\em new} closest situation from the set of expert trajectories, our method in a effect {\em adapts} to the evaluation episode.

\begin{figure}[h]
    \centering
    \includegraphics[width=0.8\columnwidth]{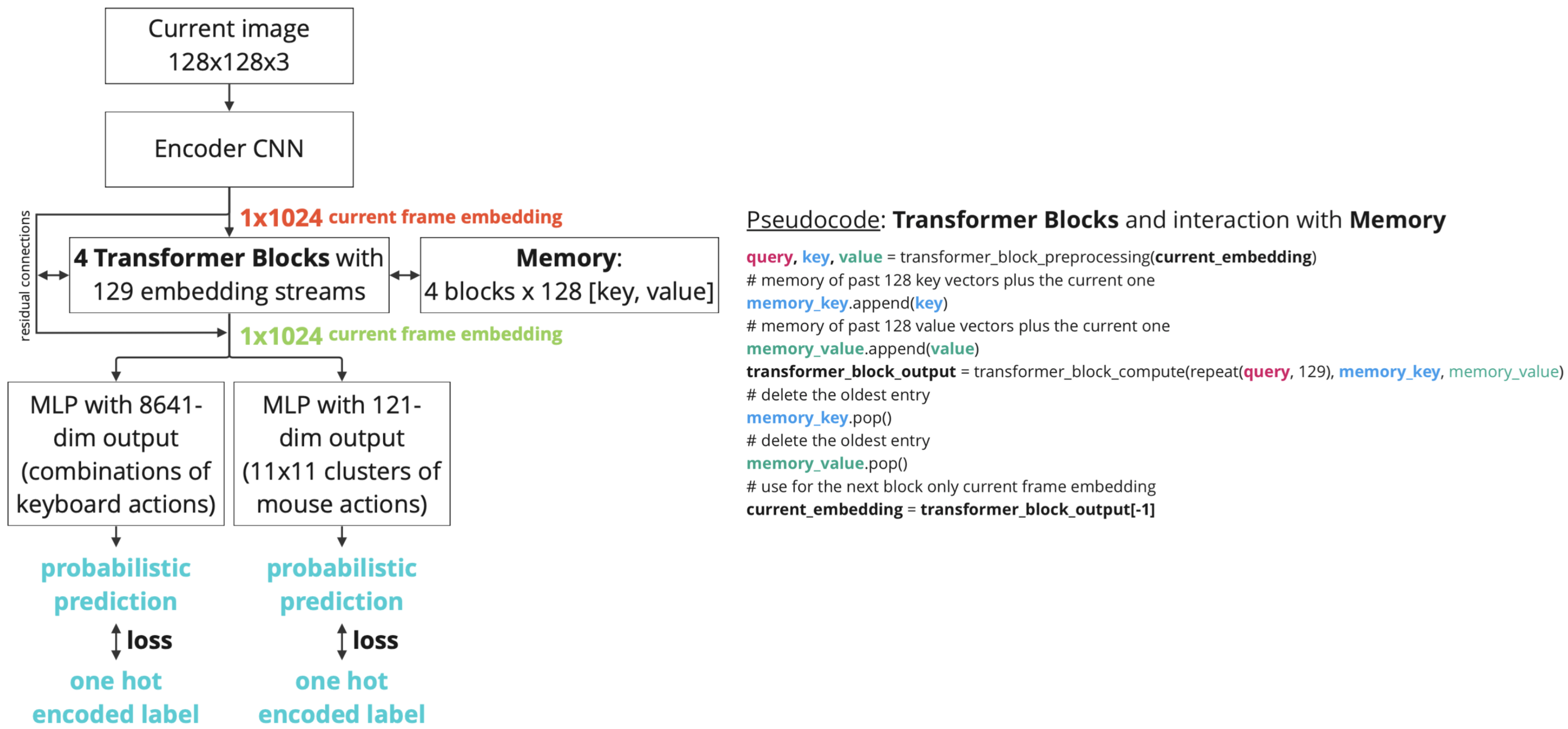}
    \caption{A scheme of the VPT model used in this study. An image input is encoded with an IMPALA CNN and passed through four transformer heads. Then, two MLP heads predict a keyboard and a mouse action respectively.}
    \label{fig:vpt_architecture}
    \vspace{-0.1in}
\end{figure}

\begin{figure}[h]
    \centering
    \includegraphics[width=0.9\columnwidth]{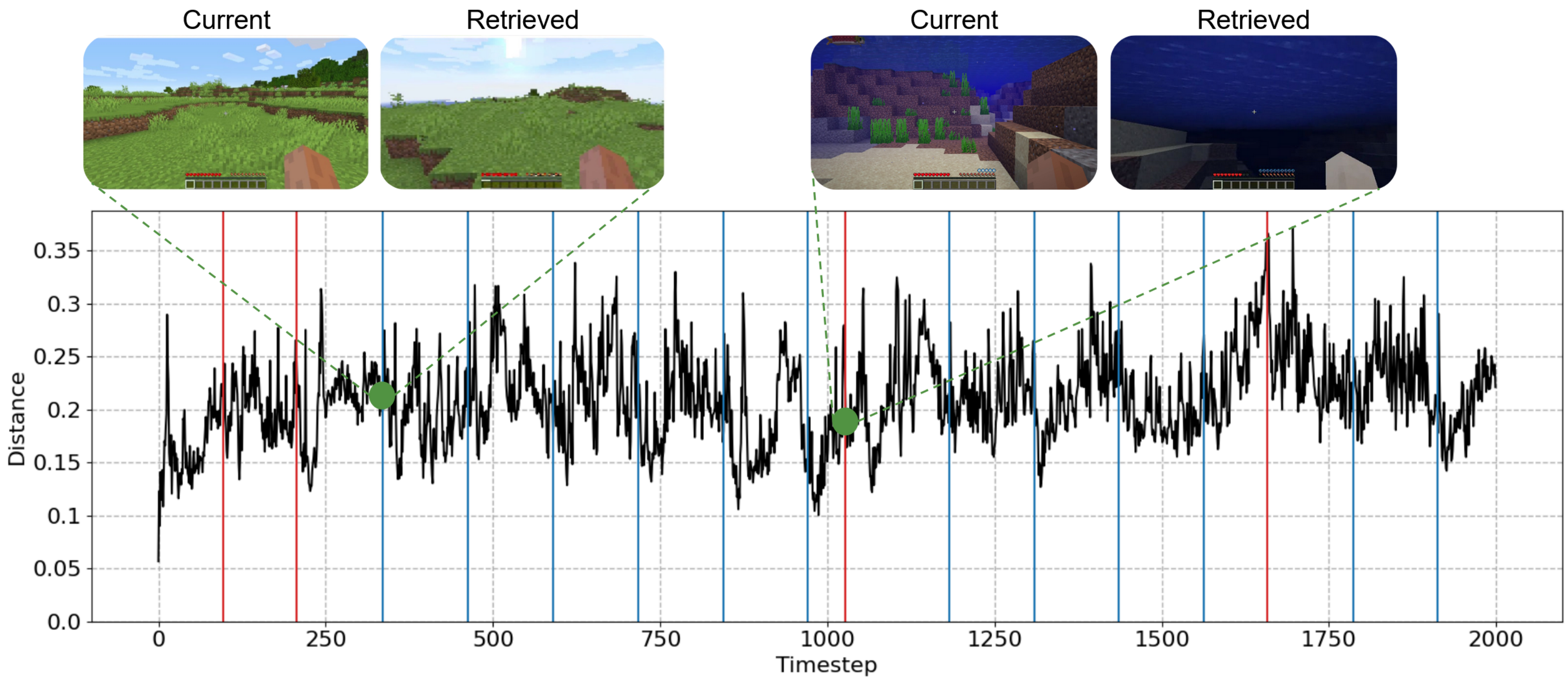}
    \caption{An example of the search mechanism. At each timestep, we keep track of the distance between current and reference embedding. Whenever the distance overcomes a threshold, a divergence-based search (red line) selects a new reference embedding; if the agent follows a threshold for too long, a time-based search (blue line) is triggered.}
    \label{fig:divergence}
\end{figure}

\section{Related work}
BC has been successfully applied to practical control problems ranging from autonomous driving~\cite{Saksena2019, Samak2021, beohar2022planning} to playing video games~\cite{KanervistoHautamaki2020, KanervistoPussinen2020, Vinyals2019}. Despite being tremendously popular due to its simplicity, BC suffers from a range of problems such as distributional shift and causal confusion~\cite{DeHaan2019}. These limitations are addressed using inverse reinforcement learning~\cite{ng2000} and generative adversarial imitation learning~\cite{ho2016generative}, which on the other hand tend to be computationally heavy for complex problems and hard to train.

Recently proposed Video Pre-Training (VPT) model \cite{baker2022video} (see Figure \ref{fig:vpt_architecture}) is a foundation model for BC trained on 70k+ hours of video content scraped from the internet~\cite{baker2022video}. 
VPT is built on an IMPALA \cite{EspeholtKavukcuoglu2018} convolutional neural network (CNN) backbone. The CNN maps an image input to a 1024-dimensional feature vector. VPT generates a batch of 129 vectors and forward them to four transformer blocks. Each block is linked to a memory block containing the last 128 frames. At the end of the transformer pipeline, only the last frame is retained and forwarded to two heads based on Multilayer Perceptrons (MLPs). One head predicts a keyboard action and second head predicts a computer mouse action. 
We provide an analysis of the architecture in Appendix A.

\section{Methods}
Our aim is to solve a complex problem in Minecraft environment, where no explicit reward is available. The only data source that we can utilize is a set of recorded expert trajectories solving that given task. The core idea behind our approach is to reformulate the control problem as a search problem on the experts' demonstrations. 

We use a pre-trained VPT model~\cite{baker2022video} to encode situations in latent space. The pre-trained version of the model used in this study is available at the official GitHub repository~\cite{VPTGithub}. The repository features three foundation models, namely \textit{1x}, \textit{2x} and \textit{3x}. The backbone of the three models is the same, and they differ only for the weights width~\cite{VPTGithub}.

\subsection{Search-based BC}

S-BC solves control problems by retrieving relevant past experiences from experts demonstrations. In this context, we define a \textit{situation} as a set of consecutive observation-action pairs $\{(o_{t}, a_{t}), \dots, (o_{t+n}, a_{t+n})\}$, $n > 0$. 

\begin{figure}[t!]
    \centering
    \includegraphics[width=0.6\columnwidth]{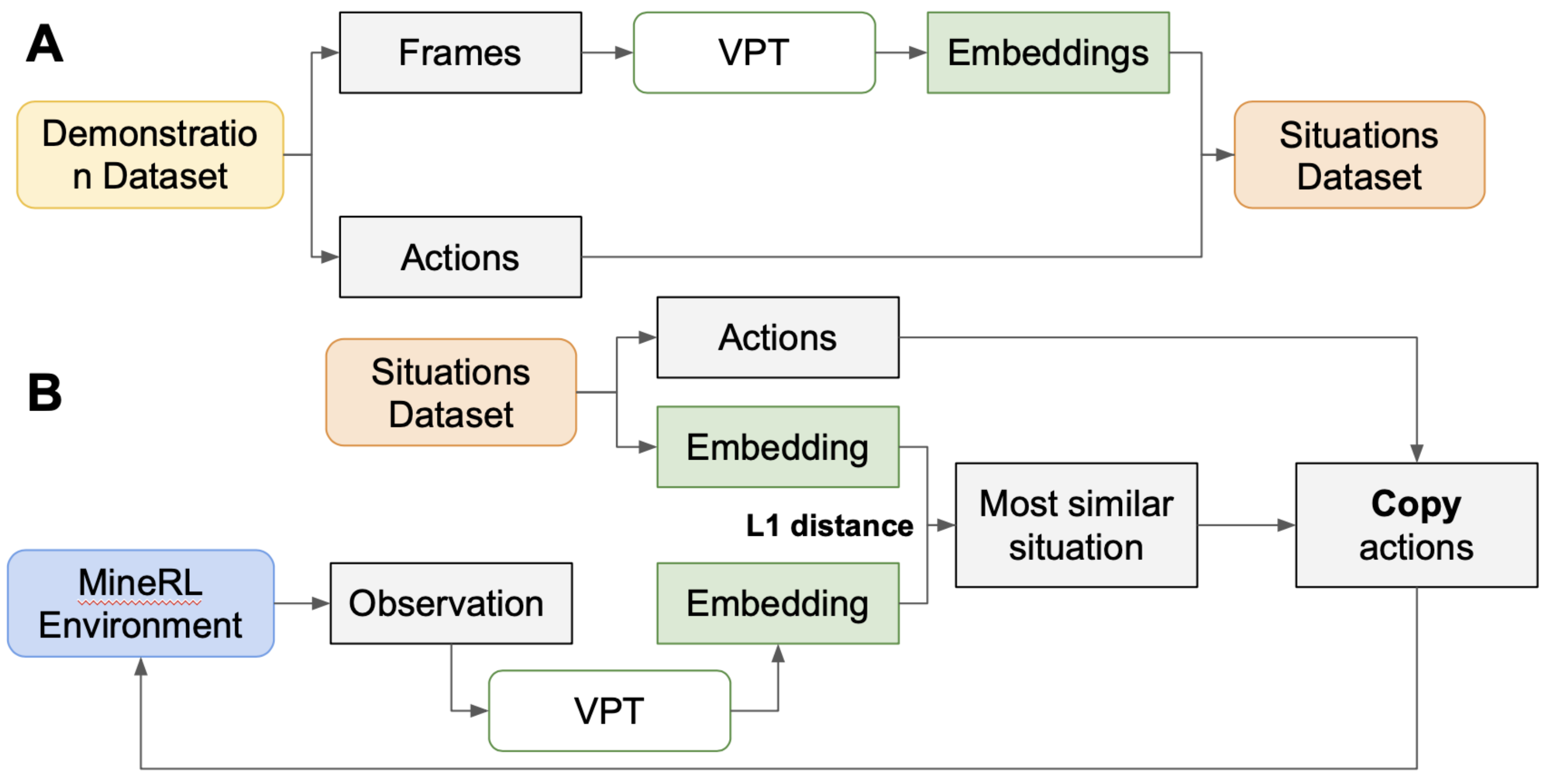}
    \caption{Our approach. (\textbf{A}) \textit{Latent space generation:} trajectories are extracted from the demonstration dataset. Frames are encoded through a provided VPT model, and paired with the corresponding actions. (\textbf{B}) \textit{Evaluation loop:} at each timestep, the new observation is forwarded to the same VPT model. Then, L1 distance across current and reference embeddings is computed and the most similar situation is found. S-BC acts in the environment following the actions of the selected reference situation.}
    \label{fig:approach}
    \vspace{-0.2in}
\end{figure}

We illustrate our approach in Figure \ref{fig:approach}. We use VPT to extract all the features embeddings from an arbitrary subset $\mathcal{S}$ of the demonstrations dataset $\mathcal{D}$, $\mathcal{S} \subseteq \mathcal{D}$. The set of these embeddings constitutes the N-dimensional latent space that S-BC searches. Moreover, the expert's optimality assumption~\cite{Russell2019} from IL methods ensures that each situation has been \textbf{addressed} and \textbf{solved} optimally.

During testing, we pass the current situation through VPT. Then, S-BC searches for the most similar embedding point in the space. Here, we compute similarity among any two situations as their L1 distance. We copy the actions of the selected reference situation. At each time-step, we shift the current and reference situations in time and recompute their distance. When situations diverge over time, the approach performs a new \textit{divergence-triggered} search (red lines in Figure \ref{fig:divergence}). If the reference embedding is followed form more than $n$ time-steps, the approach performs a new \textit{time-triggered} search (blue lines in Figure \ref{fig:divergence}).

\begin{figure}[t!]
    \centering
    \includegraphics[width=0.5\columnwidth]{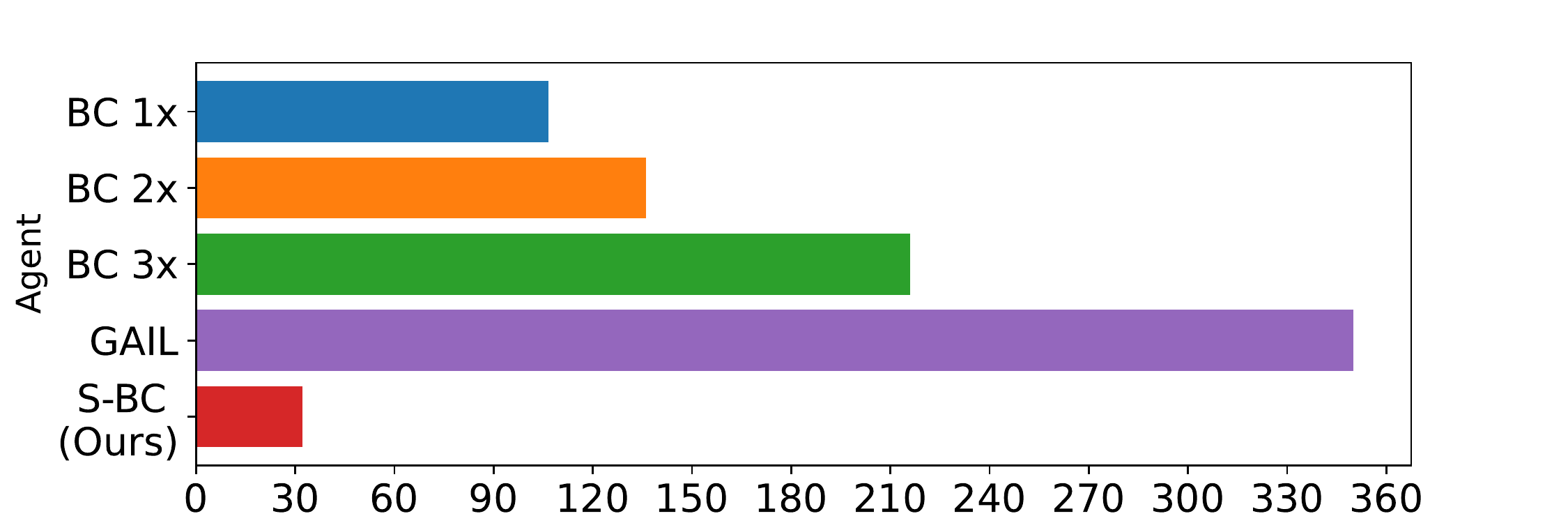}
    \caption{Time needed to train each agent on 100 trajectories on the FindCave task. In the case of BC models, the training procedure consists of fine-tuning a pre-trained VPT model. For the S-BC model, training means encoding a subset of trajectories through the reference version of VPT. All models have been trained on a single Tesla T4 GPU.}
    \label{fig:training_time}
    \vspace{-0.2in}
\end{figure}

Our S-BC process is faster than fine-tuning a VPT-based BC agent, training GAIL or RL agent (see Figure \ref{fig:training_time}). At each time-step of the evaluation (Figure \ref{fig:approach}B), the current observation coming from the MineRL environment is encoded through VPT. Then, S-BC compute the L1 distance with the current reference situation. If the two have diverged, a new search in latent space is performed and a new reference situation is selected. Otherwise, we retain the same reference situation. In both cases, S-BC copies the actions coming from the current reference embedding.

\section{Experiments}
The full FindCave dataset from the MineRL BASALT competition~\cite{ShahDragan2021} consists of 5466 experts' trajectories demonstrating the task, or around 150GB of data. For each frame of an episode, only the unprocessed RGB image and the corresponding action are available. That is, an episode is a set of image-action pairs. No reward signal is available. Similarly, no measure of performance is provided. In our study we use only a small subset of the available data. Specifically, we consider only the first $100$ trajectories.

Additionally, we personally gather expert trajectories for the MineDojo task. For all our tasks, we modified the framework to accept mouse and keyboard inputs and recorded them in the same format as the FindCave dataset. For each task, we record $100$ trajectories. Implementation details about MineDojo modifications and data gathering procedures are reported in Appendix B1.

\subsection{MineDojo tasks}
We compare S-BC against state-of-the-art models in the Minecraft (VPT-based $1x$, $2x$, and $3x$ models) and IRL domains (GAIL). We fine-tune each VPT-based model for two full epochs on our gathered data, and fine-tune a different model for each task. Moreover, we train GAIL using VPT-encoded trajectories, to speed up training by reduce the size of the observation space. We train GAIL for ~6h for each task. We evaluate the models in two ways. First, we use the MineDojo \cite{fan2022minedojo} framework to provide clear numerical results. 

In the suite, authors distinguish \textit{ground-truth} and \textit{creative} tasks. On one hand, ground-truth tasks feature a well-defined terminal condition, such as "kill a specific unit" or "gather $x$ quantity of material", and are separates in broad categories such as \textit{combat}, \textit{build}, \textit{harvest}, and so on. On the other hand, creative tasks do not feature a terminal condition. Although, they can be evaluated indirectly through MineCLIP \cite{fan2022minedojo} by mapping a text prompt and a video frame in a shared latent space and then computing cosine distance to assess similarity between them. Such similarity score is computed for each timestep, as in standard RL. 

For our experiments, we select four ground-truth tasks, namely, \textbf{combat spider} and \textbf{combat zombie} from the \textit{combat} category and \textbf{harvest log}, \textbf{harvest wheat seed} from the \textit{harvest} category, and two \textit{creative} tasks, \textbf{"find ocean"} and \textbf{"dig a hole"}. 

\subsection{FindCave}
As second set of experiments, we compare the agents directly on the FindCave task from BASALT \cite{Milani2023}, where no reward information is known or obtainable. To support the evaluation process, we train a simple binary classifier to detect the presence of a cave. Our classifier is composed by four convolutional layers and two fully-connected layers. The network has been trained with supervised learning on a dataset of cave frames, extracted from the original FindCave data, and achieved a validation accuracy of $97.89\%$ over it. Further details are discussed in Appendix B2. To ensure fairness over the amount of accessible relevant knowledge to each model, we use the same subset of $100$ trajectories from the original BASALT dataset on them.

We apply minimal changes to the task rules with respect to the official competition, to ease the evaluation procedure while keeping the performance evaluation intact. First, we disable the terminal action \textit{'ESC'} in all the agents. We support this decision by highlighting that terminal actions constitute a minimal fraction of the amount of training data, and a BC agent would likely ignore it, or use it incorrectly. Therefore, we removed the action to evaluate the Second, the BASALT competition considers an episode successful whenever an agent performs the terminal action while being inside a cave. Instead, we consider an episode successful whenever an agent spends more than five seconds in a cave. The maximum time limit for each evaluation episode is set to three minutes, similarly to the official competition rules.

As in the evaluation process of the BASALT competition, we test our agent on twenty seed values. Since the official values used in BASALT evaluation are not publicly available, we have selected each seed manually to ensure the presence of caves. For each run, we play 10 evaluation episodes.

\paragraph{Latent space visualization}
We perform a study over the latent space used from our agent. We plot t-SNE representation of a full latent space and evaluate position of the cave frames, similarity between close points. Finally, we track the movement of an agent using such latent space, to study its dynamics.

\paragraph{Evaluation procedure}
Our evaluation procedure is as follows. First, we let an agent play the evaluation episodes and record each one of them. Then, we pass each frame of the video to the classifier, and label them as either \textit{cave} or \textit{non-cave} frames. Whenever a cave frame is detected, we start counting the number of consecutive cave frames. Following the MineRL specification \cite{GussSalakhutdinov2019}, we assign each frame a duration of $0.05$s. This lead to set the threshold of five seconds to $100$ consecutive cave frames. Whenever such threshold is exceeded, we consider an episode successful.

To account for unique situations, we run additional evaluations lowering the time threshold. For each spotted situation, we take a look at the current frame and evaluate it manually. Unique situations include, but are not limited to, agent reaching a cave and being killed before the five second mark, agent digging a cave itself (not allowed from BASALT competition rules), or agent walking in exceptionally dark areas that might be mistaken as caves.

\section{Results}

\begin{table*}
\centering
{\addtolength{\tabcolsep}{0pt}
\begin{tabular}{c c c c c c}
  & \textbf{BC 1x} & \textbf{BC 2x}  & \textbf{BC 3x} & \textbf{GAIL} & \textbf{S-BC (Ours)} \\[2pt] \hline \\[-7pt]
Find Ocean & $20.25\pm1.10$ & $20.19\pm1.04$ & $19.82\pm1.05$ & $20.30\pm0.74$ & $\textbf{20.47}\pm1.11$\\[2pt] \hline \\[-7pt]
Dig Hole                               & $20.80\pm0.88$                              & $20.94\pm0.85$     & $20.95\pm0.87$ & 
$20.20\pm0.24$ &  $\textbf{20.97}\pm0.92$                               \\[2pt] %
\end{tabular}
\caption{Average MineCLIP score and its standard deviation per model on the evaluated \textit{creative} tasks. MineCLIP scores can be used to train agents in the MineDojo suite in an RL fashion.}
}
\label{tab:mineclipscores}
\end{table*}

\begin{figure}[t!]
    \centering
    \includegraphics[width=0.8\columnwidth]{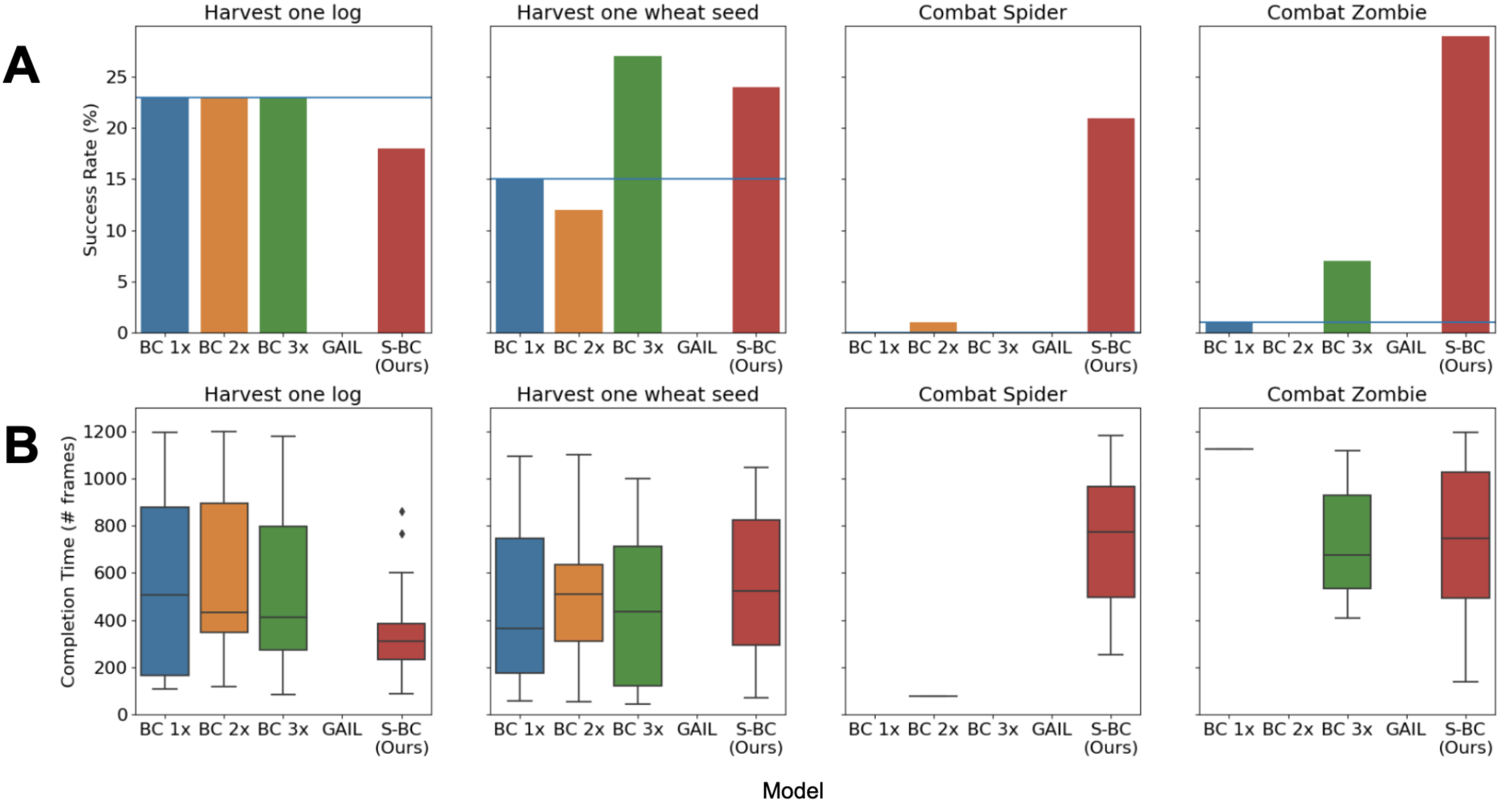}
    \vspace{-0.1in}
    \caption{Results for four MineDojo ground-truth tasks. \textbf{(A)} Model success rate over 100 evaluation episodes. Reference baseline model BC 1x is highlighted with a horizontal blue line in each plot. \textbf{(B)} Average completion time for each task. Tasks that have never been completed are associated with a blank space. Tasks completed only one time show a single line corresponding to the completion time of that single success.}
    \label{fig:dojo_results}
    \vspace{-0.2in}
\end{figure}

\subsection{MineDojo tasks}

We evaluate our models over six diverse tasks of the MineDojo suite. Figure \ref{fig:dojo_results} shows the models success rate and completion time for each task. Additionally, we report in Table \ref{tab:mineclipscores} the average score obtained by each model in each creative task.

Notably, S-BC performs similarly or better than its competitors. Figure \ref{fig:dojo_results}A shows that our method is the only one able to complete instances of the combat tasks. In harvest tasks, our model either performs similarly or lose a mere 5\% in performance with respect to much bigger, learning-based models. It is also interesting to notice that VPT-based foundation model were originally trained on the ObtainDiamond track~\cite{kanervisto2022minerl}, which first step consists in collecting logs. We highlight that GAIL was not able to complete any of the ground-truth tasks. We point out that we have trained GAIL for a time that is comparable to the fine-tuning time of the VPT-based models (~1h). Therefore, due to the limited amount of time used for training, it might under perform on some tasks.

Figure \ref{fig:dojo_results}B shows the average time of completion for successful episodes. We highlight that S-BC requires either the same on less time to complete, on average. Our agent is on average the faster in completing \textbf{harvest one log}, and performs similarly in \textbf{harvest one wheat seed}. In the \textbf{combat spider} case, S-BC was the only method able to complete the task consistently, using on average two thirds of the time at its disposal.

Table \ref{tab:mineclipscores} describes the results of the \textbf{creative} tasks. According to MineCLIP, all the models performed similarly. It is interesting to notice that in creative tasks, GAIL seems to achieve comparable performance to other methods.

\begin{figure*}[!t]
    \centering
    \includegraphics[width=0.5\textwidth]{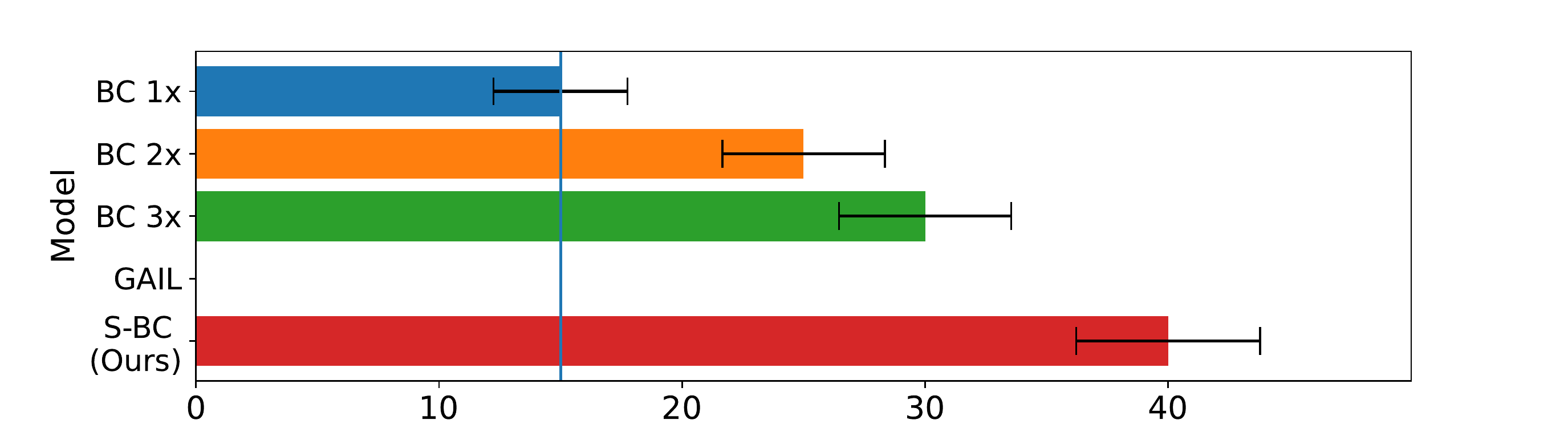}
    \caption{Average success rate for our tested model on the FindCave task over $3$ runs. Each agent has been evaluated over $20$ seeds, playing 10 episodes on each seed. Baseline model is highlithed with a vertical blue line.}
    \label{fig:cave_results}
    \vspace{-0.2in}
\end{figure*}

\subsection{FindCave}
We report success rate for the models on the FindCave task in Figure \ref{fig:cave_results}. S-BC has obtained the best performance, being able to complete the task 40\% of the times. GAIL was never able to find a cave, while VPT-based BC models were quite successful, reporting success rates of $15\%$, $25\%$ and $30\%$ for the \textit{1x}, \textit{2x}, and \textit{3x}. 

\begin{figure*}[!t]
    \centering
    \includegraphics[width=0.95\textwidth]{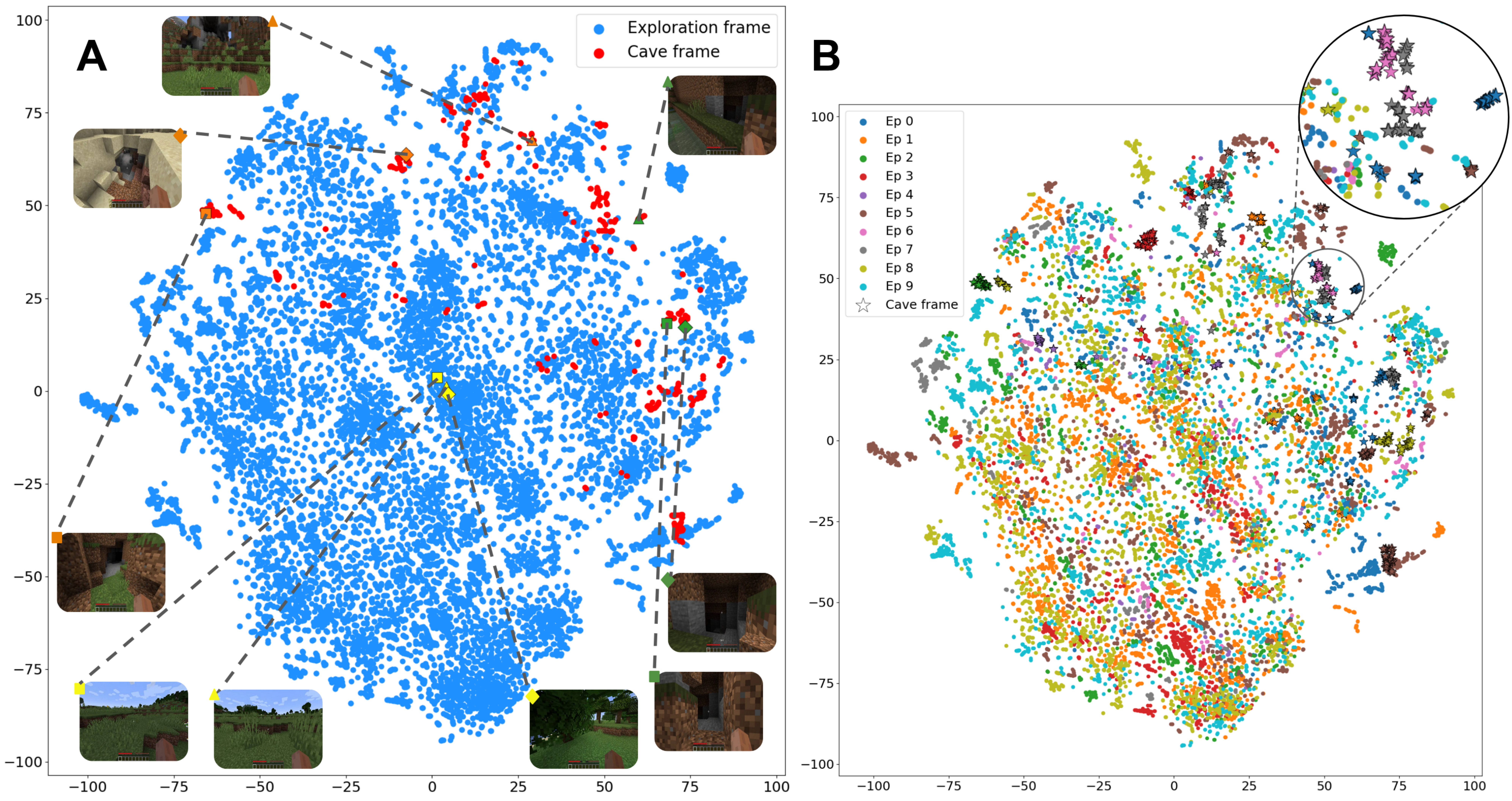}
    \caption{t-SNE diagrams of S-BC latent space with $10$ trajectories. \textbf{(A)} Separation between exploration (blue) and cave (red) frames. Some examples of cave frames are linked to the corresponding point (symbols) in the latent space. \textbf{(B)} t-SNE representation of the same space with points separated according to their belonging trajectory. Cave frames for each trajectory are marked with a matching color star symbol. \textit{Zoom:} a close-up look at some clustered cave frames coming from different trajectories.}
    \label{fig:latent_plot}
    \vspace{-0.2in}
\end{figure*}

\subsection{Latent space visualization}
We have analyzed the VPT latent space used by our models and its transformation over the number of encoded trajectories. To visualized it, we have used t-SNE \cite{hinton2008} plotting. Figures~\ref{fig:latent_plot}A and~\ref{fig:used_trajectories} show two examples of latent space used by S-BC, with $10$ and $25$ encoded trajectories respectively. We have classified each point as either a cave (red) or exploration frame (blue). We used our cave classifier to determine all cave frames. In our study, we have analyzed visual hints of structure of those spaces. Additionally, we have extracted some meaningful pieces of information such as similarity across frames and the search path of an episode.

\subsubsection{Space structure \& cave frames}
Both $10$ and $25$ trajectories latent space plots (Figure \ref{fig:latent_plot}A and \ref{fig:used_trajectories}, respectively) show that the embeddings in the VPT latent space can be separated to some degree, but are quite dense. Perhaps not surprisingly, an increase in the number of encoded trajectories results in a less structured space. This is supported by the visual appearance of a Minecraft frame: the game features several biomes, that range from desert to snowy mountains. In the FindCave task, an agent is bounded to spawn in a plain biome. Such biome is characterized by the presence of fields, forests, ponds and, sometimes, lakes or sea. Therefore, a number of visually different frames are possible. But, whenever switching from two different environments (e.g. from lake to forest), we expect some "transition frames" that contain a mixture of elements of both. In a visualization method such as t-SNE, we expect such frames to be equally distant from proper examples of any of the two environments. That is, we expect them to be in the middle of two more dense regions of space. So, as the number of trajectories increase, we might expect the number of transition frames to increase, thus making well-defined clusters more rare.

The position of cave frames within the space constitute another notable result of this visualization. The vast majority of cave frames are grouped in half of the space. Within this region, some clusters are easily spotted, but in general cave frames tend to be sparse.

\subsubsection{Visual similarity across points}
In our study, we have visualized several frames of the latent space, to understand the relation between their visual resemblance and spatial coordinates. We show some examples in Figure~\ref{fig:latent_plot}A: frames linked with a \textit{green} symbol are cave frames belonging to the same trajectory; frames corresponding to an \textit{orange} symbol are cave frames coming from different trajectories; finally, frames marked with a yellow symbol are exploration frames belonging to the same cluster. 

By comparing green and orange examples, it is clear that each cave cluster contains frames from a different trajectory. Additionally, frames linked to a yellow symbol show how each cluster is formed by visually similar examples. Such morphology of the space suggests that, depending on the current situation, an S-BC agent will select a close point whenever the visual transition across situations is smooth. On the contrary, whenever the observation drastically changes, the agent will look for a distant new situation to follow. 

To account for the points distribution of each trajectory, we have plotted a second version of the $10$ trajectories latent space (Figure \ref{fig:latent_plot}B). In this representation, each point is colored according to its belonging trajectory. We have marked the cave frames of a specific trajectory with a star symbol of the same color. 

Notably, each cluster contains points of a variety of colors. This supports the fact that visual resemblance plays a fundamental role in deciding a point position within the space: similar frames (e.g. fields, forests, and so on) coming from different trajectories are squeezed together. Therefore, one might conclude that VPT embeddings are heavily biased towards visual similarity. While we believe that such metric is indeed important, we have identified clues that suggest other important factors in defining them. In particular, we believe that past information is another key factor in this regard. To support this claim, consider the yellow examples: while the square and triangle seems to be the most visually similar pair, their distance is greater than e.g. triangle and rhombus. Such observation imposes to consider other factors in the characterization of a VPT embedding.

\subsubsection{Example of one episode using S-BC}
\begin{figure}[t!]
    \centering
    \includegraphics[width=0.5\columnwidth]{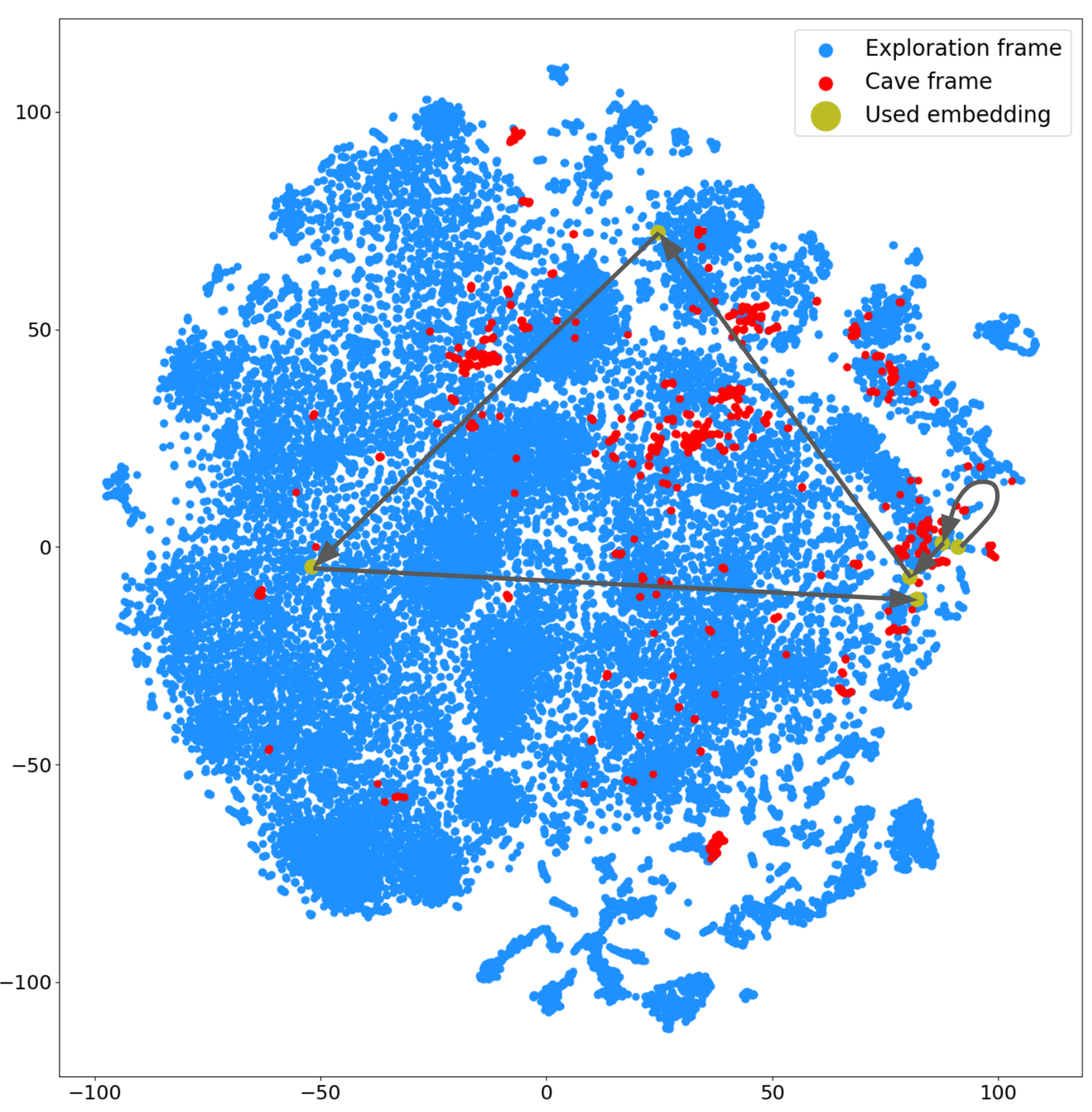}
    \caption{Visualization of the VPT embeddings latent space with 25 trajectories encoded. A yellow point marks a trajectory used from the agent during the corresponding episode.}
    \label{fig:used_trajectories}
    \vspace{-0.2in}
\end{figure}
We have visualized the situations that an S-BC agent trained on 25 trajectories has followed in a successful evaluation episode. The result is shown in Figure~\ref{fig:used_trajectories}. Each yellow dot corresponds to a new search performed by our agent at some point in the episode. Notably, $4$ out of $6$ searches were performed in a cave frames cluster. In the episode, the agent enters the cave after around $15$s. Then, it wanders there and gets killed after $30$s. The fact that our agent has performed a total of $6$ searches suggests that, on average, it has followed each selected trajectory for $5$s (around $100$ frames).

\subsection{Perceptual evaluation}

\begin{table*}[!t]
\centering
{\begin{tabular}{c c c c c c}
\hline \\[-6pt]
\textbf{Team} & FindCave & MakeWaterfall & AnimalPen & House & \textbf{Average} \\[1pt]
 \hline \\[-6pt]
GoUp & $0.31$ & $\textbf{1.21}$ & \textbf{0.28} & \textbf{1.11} & \textbf{0.73} \\[1pt]
\textbf{Our approach} & \textbf{0.56} & -0.10 & 0.02 & 0.04 & \textbf{0.13} \\ [1pt]
voggite & 0.21 & 0.43 & -0.20 & -0.18 & \textbf{0.06} \\[1pt]
JustATry & -0.31 & -0.02 & -0.15 & -0.14 & \textbf{-0.15} \\[1pt]
TheRealMiners & 0.07 & -0.03 & -0.28 & -0.38 & \textbf{-0.16} \\[1pt]
\hline \\[-6pt]
Human2 & 2.52 & 2.42 & 2.46 & 2.34 & \textbf{2.43} \\[1pt]
Human1 & 1.94 & 1.94 & 2.52 & 2.28 & \textbf{2.17} \\[1pt] \hline \\[-6pt]
BC-Baseline & -0.43 & -0.23 & -0.19 & -0.42 & \textbf{-0.32} \\[1pt]
Random & -1.80 & -1.29 & -1.14 & -1.16 & \textbf{-1.35} \\[1pt] \hline \\[-9pt]
\end{tabular}
\caption{Top-5 ranking of the NeurIPS BASALT 2022 competition~\protect\cite{Milani2023}. Below, the TrueSkill~\protect\cite{minka2018trueskill} scores for two human expert players, a BC baseline and a random agent are reported for comparison.}
}
\label{tab:results}
\vspace{-0.2in}
\end{table*}

To determine its rank among the participants, the organizers of the competition have used the TrueSkill~\cite{minka2018trueskill} ranking system, which is widely used in the Microsoft online gaming ecosystem. Given a set of competitors, the system uses Bayesian inference to compute an ELO-like score, according to the match history of each competitor.

In the context of the BASALT competition, the system considers each trained agent as a participant. To create a match history for each of them, any two participants are drawn and their evaluation videos are judged by a human contractor. For each pair of videos, the contractor indicates which agent was able to solve the task (or which one was closer), and how human-like an agent appeared to be. Those two ratings are then aggregated to the match history of each competitor. When a suitable amount of outcomes are available for an agent, the TrueSkill for that agent is computed.

Our proposed S-BC agent was overall ranked second place in the challenge. The results of our agent are described in \ref{tab:results}. First place was awarded to team GoUp. Their method uses detection methods and leverages human knowledge of the task through scripting \cite{Milani2023}. Notably, all the other learning-based methods achieved lower performance than S-BC. Additionally, our method was awarded with $2$ out of $5$ research innovation prizes.

\subsubsection{Behavior of agents in videos}
In our study we have collected and analyzed a wide number of videos that demonstrate the general behavior of each agent. From those, we were able to delineate some differences in the behavior of BC and S-BC agents:
\paragraph{BC models}
As a premise, it is important to note that the reference version of VPT provided by the organizers of the competition was trained on data from the MineRL ObtainDiamond track~\cite{kanervisto2022minerl} and then fine-tuned on the BASALT~\cite{ShahDragan2021, Milani2023} dataset. Judging from the videos, our additional fine-tuning towards the FindCave task shows some improvements of the agents. Overall, though, the general behavior remains the primary objective of the ObtainDiamond competition: it is not rare that a BC agent chops trees as first step. We believe that the limited amount of episodes used for fine-tuning has had a clear negative impact on the performance. We expect an agent fine-tuned on the full FindCave dataset to be more aligned with the expected FindCave behavior, at the cost of a drastic increase in training time.
\paragraph{S-BC}
We found our agent to be quite robust in performance, and be human-like at the same time. S-BC shows significantly more aligned behavior with the proposed task than any BC agent. In general, our agent is able to recover meaningful trajectories when the cave is nearby. Despite this, our agent is far from being perfect: in some instances, S-BC has shown dichotomic behavior regarding caves. At first, the agent would face a cave and quickly turn around. In the following seconds, a new search correctly guiding the agent towards the same cave was performed. We believe that this behavior comes from the possible presence of enemies in caves, which would influence the retrieval of a better suited reference situation.

\section{Conclusions}
We introduced S-BC, a BC-inspired search-based approach that uses the past experience of experts to solve similar control problems. Our approach shows better or similar performance, requiring less training time, allowing a few-shot learning, and multi-skill performance when skill demonstration is provided. We applied our method to the MineRL BASALT NeurIPS 2022 Challenge, where our agent reached overall 2nd place. 
The agent had to demonstrate human-like behavior while completing the tasks. Numerical evaluations over a suite of MineDojo tasks shows that our agent reaches similar or better performance than state-of-the-art BC and IRL methods, while requiring no re-training or fine-tuning.

\bibliography{bibliography.bib}
\bibliographystyle{ieeetr}

\newpage
\appendix

\section{Appendix - Architectures details}
\subsection{Details on VPT architecture}

OpenAI's Video Pre-Training models(6) were trained using behavioral cloning on a large video dataset, resulting in the \textit{foundation} models that were able to solve simple tasks in the MineRL environment without further fine-tuning like RL. The authors also provide fine-tuned model variants, which are more specialized to certain tasks. For similarity search of the embedded situations and experiments, we used a pre-trained \textit{foundation-model-1x.weights}\footnote{https://openaipublic.blob.core.windows.net/minecraft-rl/models/foundation-model-1x.weights} VPT model. 

The VPT models take a single RGB frame, resized to $128\times128$, as the input, which is passed through an IMPALA CNN (25) to generate $1024$-dimensional image embeddings. These embeddings are then passed to four residual, recurrent transformer blocks. These transformer blocks contain a memory state, which consists of \textit{[key, value]} pairs of the past $128$ steps (see Figure \ref{fig:vpt_architecture_appendix}), but only use the current input as the query. After these blocks, a $1024$-dimensional intermediate representation is put out before the MLP heads, which are used to predict keyboard and computer mouse actions. These intermediate representation latents were used for the embedding of situations. OpenAI's VPT code implementation was made publicly available on GitHub (18).

\renewcommand{\thefigure}{9}
\begin{figure}[h]
    \centering
    \includegraphics[width=0.8\columnwidth]
    {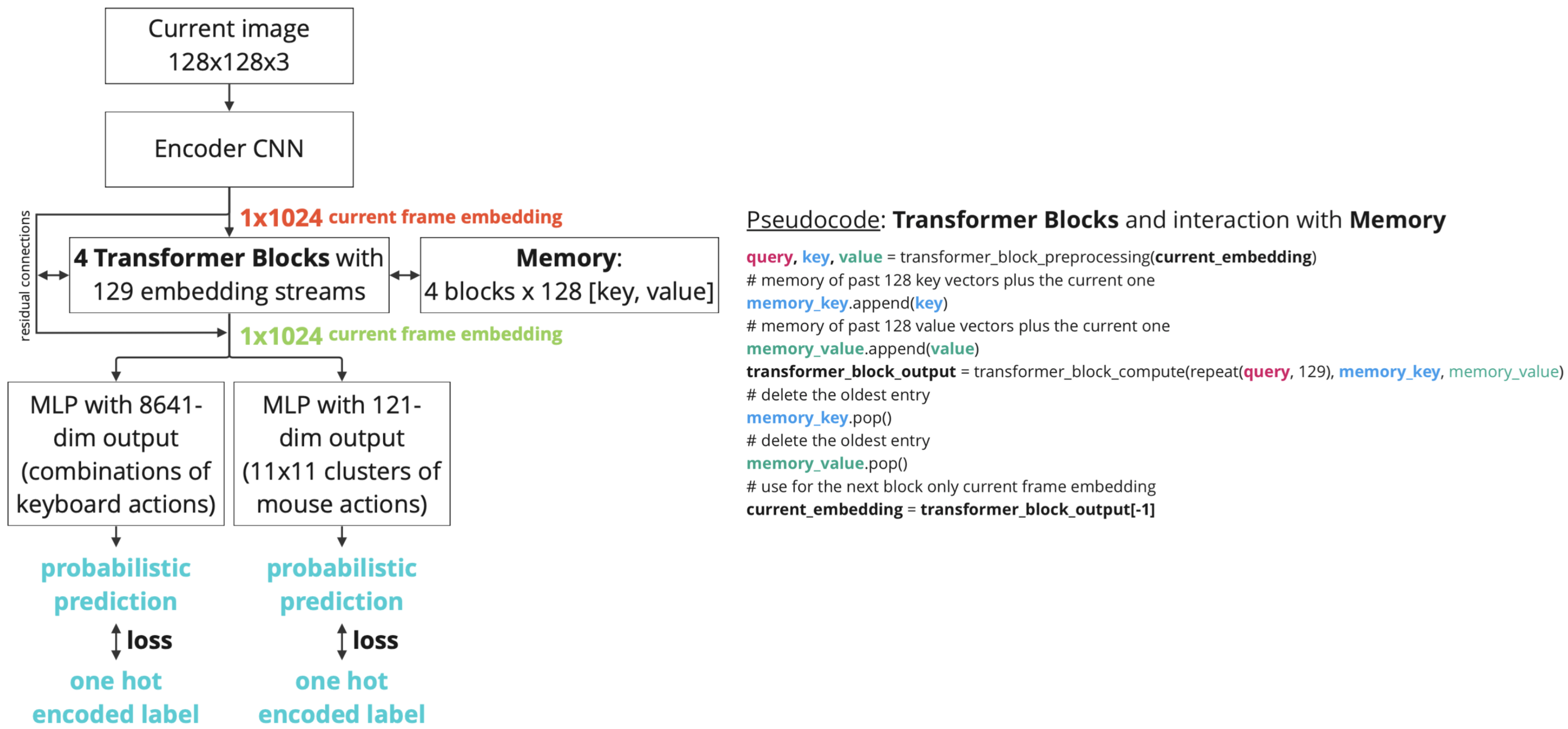}
    \caption{A scheme of the VPT model used in this study. An image input is encoded with an IMPALA CNN (25) and passed through four transformer blocks. Then, two MLP heads predict a keyboard and a mouse action respectively.}
    \label{fig:vpt_architecture_appendix}
    \vspace{-0.1in}
\end{figure}

\subsection{Details on MineCLIP}
\renewcommand{\thefigure}{10}
\begin{figure}[h]
    \centering
    \includegraphics[width=\columnwidth]{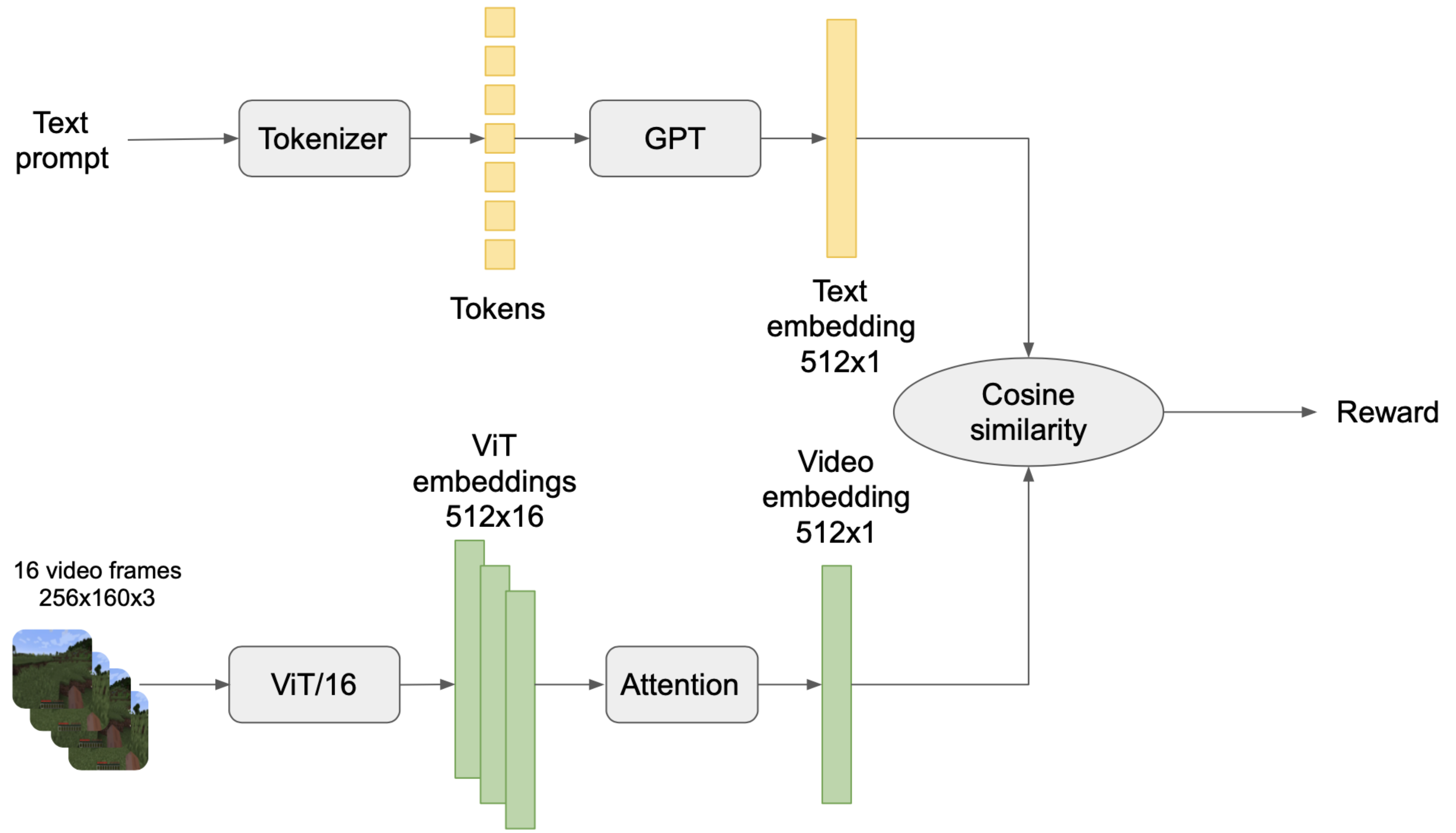}
    \caption{A high level scheme of the MineCLIP architecture. On one branch, a text prompt is encoded with a variant of GPT on a 512d embedding. On the other branch, 16 video frames are encoded through ViT/16 in an embedding of the same size. Finally, a reward is computed using cosine similarity among these two embeddings.}
    \label{fig:mineclip_arch}
    \vspace{-0.1in}
\end{figure}

We use a pretrained MineCLIP (20) model for the evaluations, provided by the original authors. Its architecture, as seen in Figure \ref{fig:mineclip_arch}, consists of three main components: an image encoder, a temporal aggregator and a text encoder. The output scores are computed via a cosine similarity measurement between the encoded observations and the encoded text prompt. 

\begin{itemize}
    \item The frame-wise image encoder utilizes the ViT-B/16 architecture (26) and embeds input images of size $256\times160$ into a $512$-dimensional latent space.
    \item The temporal aggregator originally comes in two variants, with one being based on average pooling, and the other utilizing self-attention. We used the attention variant, as it had better results in the original paper (20). It is a 2-layer, self-attention transformer with an embedding size of $512$ and $8$ attention heads. The authors also added two additional residual CLIP Adapter (27) layers after the aggregator.
    \item The text encoder utilizes a $12$-layer GPT with $8$ attention heads and an embedding size of $512$. The corresponding text tokenizer has a maximum length of $77$ tokens and uses a vocabulary of size $49152$.
\end{itemize}

MineCLIP is trained to assign a score to each frame based on the cosine similarity between a text and an image embeddings, consistently with the classic RL paradigm. Therefore, using the average episode score is a coherent choice with assessing the quality of an agent.

\paragraph{Evaluation procedure}
During the evaluation, we resize the observed frames from the original VPT resolution (6) of $640\times360$ to the appropriate MineCLIP size of $256\times160$, and use a sliding window of $16$ consecutive frames as the input, as was proposed in the original paper (20).
To compute the final MineCLIP scores per task, we used the text prompts specified by MineDojo: "Explore the world to find ocean." for the \textbf{find ocean} task and "Dig a hole in the ground." for the \textbf{dig a hole} tasks.
\section{Appendix - Implementation details}
\subsection{MineDojo}

For our experiments, we performed modifications on MineDojo. To be able to use the OpenAI VPT models, which were created for MineRL 1.0 while MineDojo is based on MineRL 0.4.4, the action space of the environments had to be adapted.

This includes removing event-level control, such as equip or craft actions, and adding hotbar selection actions and fine-grained mouse control. These modifications also allowed us to create a human interface to collect expert trajectories for the corresponding tasks.

Additionally, we enabled the "fast reset" option for all environments, which significantly speeds up their reset times by teleporting the agent to a random place far away on a reset instead of generating an entirely new world each time. However, by specifying a large number as the range parameter for this option, diversity is still ensured.

More specifically, we used the following parameters for the different MineDojo environments:

\begin{itemize}
    \item High range for fast resets of $5000$ blocks, $2000$ blocks for the \textbf{combat zombie} and \textbf{combat spider} tasks.
    \item Timelimit of $1200$ steps for all environments, i.e., $1$ minute real-time at $20$ steps per second.
    \item RGB observation space with size $640\times360$.
\end{itemize}

\renewcommand{\thefigure}{11}
\begin{figure}
    \centering
    \includegraphics[width=0.9\columnwidth]{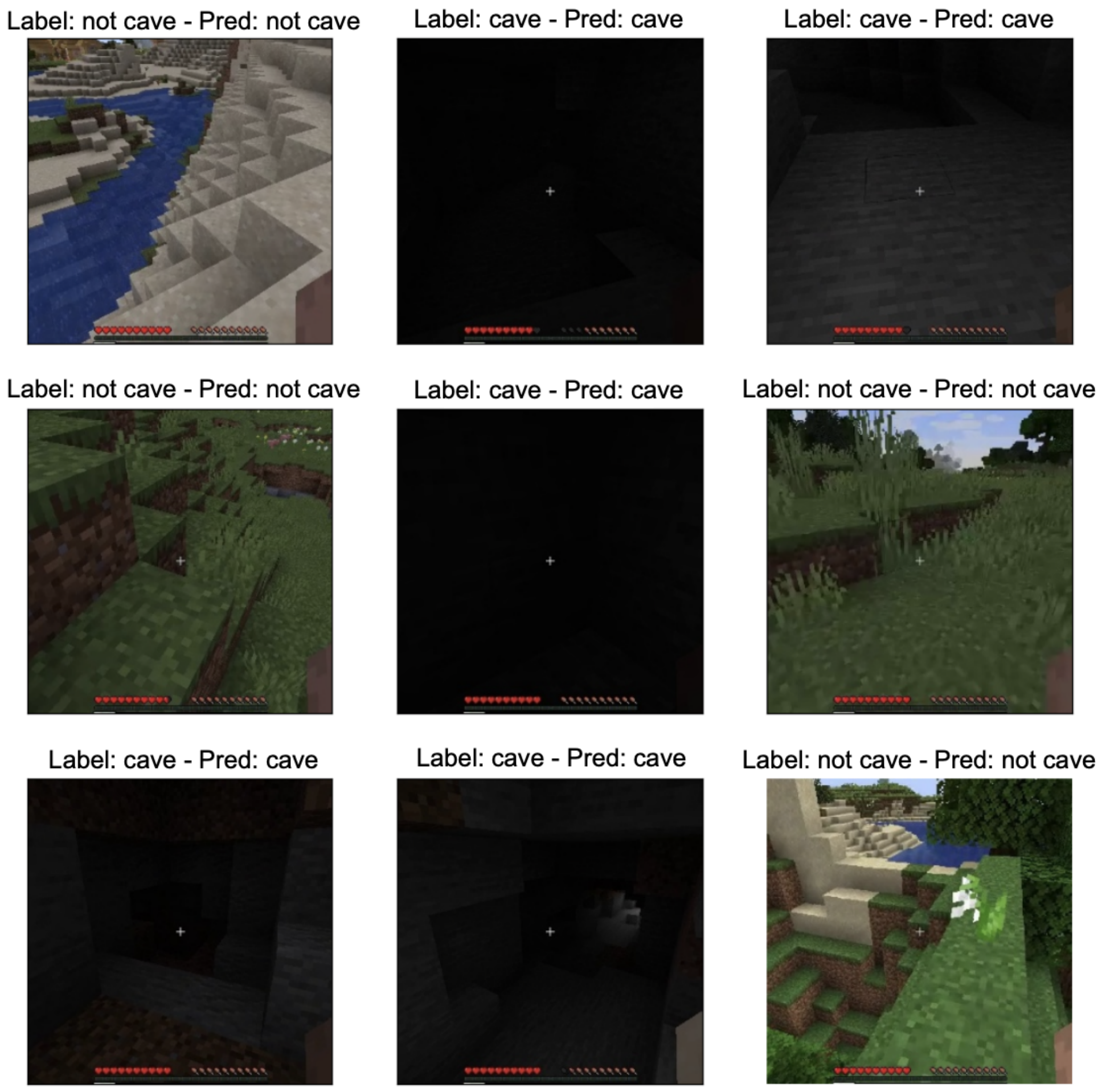}
    \caption{Some examples of cave classification. A cave is spotted when an agent stands in front of it (third row, first image) and in generally dark frames (second row, second image; third row, third image).}
    \label{fig:cave_class}
\end{figure}

\subsection{Cave classifier}

\paragraph{Dataset creation} 
The dataset for the cave classifier has been created from the original FindCave dataset. Such dataset features 5466 demonstrations from experts solving the task. For each of them, we use a combination of heuristics and manual relabeling to ensure data coherence. First, it is worth
noting that a cave will appear only in the ending part of an episode. We estimate that a cave frame will appear around the last 2 seconds of a successful episode (that is, within the last 40 frames of an episode). Therefore, we empirically select the $30^{th}$-to-last frame from each trajectory and extract it. This ensured us to have around 5000 cave frames, but also some spurious frames coming from unsuccessful episodes.

We removed a number of spurious frames by computing the average pixels value in the 320 × 320 centered region of each image. We motivate this choice by noting that a real cave frame would probably feature dark pixels at least in its central region. Still, some specific situations (e.g. a particularly dark forest) would resist to this method. We removed those frame by manually going through the data. 

To account for the \textit{non-cave} class examples, we randomly extracted one frame from the first half of each episode. As the FindCave dataset is composed by trajectories from human experts, the presence of a cave in a frame is ultimately decided by human judgement. Therefore, there is a chance of having cave frames among the examples of the non-cave class. To address this problem, similarly to the opposite class, we compute the average pixel values on the central region of the image, and remove the false negatives manually.

Finally, we added the spurious frames of each class to the correct one. We removed only the ones that were showing exceptional situations e.g. open inventory, or ending screen of an episode. In the end, we counted 5456 cave and 5455 non-cave examples. Each example has been resized to match the input dimension of 320 × 320 by central cropping.

\paragraph{Cave classifier architecture} The architecture of our cave classifier is made of four convolutional layers, each with 64 channels, 3 × 3 kernels, and ReLU activation. The input size is a 320 × 320
RGB image. Then, we flatten the embedding and forward it to two fully-connected layers of 1024 units each, with ReLU activations. At the end of the network, we apply softmax activation layer to get a probability distribution over the two output classes.

\paragraph{Training procedure} We trained our cave classifier using a 90/10 splitting for training and validation data. We have trained the network for 10 epochs, until convergence. We tested our classifier over a subset of the validation data and obtained an accuracy of 97.89\%. Figure \ref{fig:cave_class} shows some classification examples.
\section{Appendix - Evaluation details}
Evaluating a model intended to work "like a human" is hard, as to this day no comprehensive metrics for assessing human-likeliness is known. As an attempt to encapsulate this complex set of behaviors in a number, BASALT organizers use the TrueSkill score, as explained in Section 5.4. According to this metric, our agent has indeed shown human-likeliness to some extent.

Nonetheless, assessing the human-likeliness of an agent does not answer questions about its capabilities and limitations. In this regard, TrueSkill is indeed limited: first, the score aims to establish a global ranking among $n$ agents competing against each other. Second, in the scope of BASALT such ranking is based on a number of pairwise human evaluations, which is time consuming to obtain. To gather an in-depth view of our agent capabilities, we require a  reliable and automatically computed metric.

Therefore, success rate represents the only objective way of evaluating the agents of this study. Differently from the official BASALT competition, in the FindCave context we define an episode to be successful whenever the agent enters a cave and stays there for more than five seconds. We base our decision on two main facts:
\begin{itemize}
    \item By definition of the task, a successful agent whose scope is to reach a cave, will stay there indefinitely once it enters one (that is, if it works as intended);
    \item Any agent that uses some past reference, will need to demonstrate that the action of entering the cave was willingful.
\end{itemize}
We argue that, in addressing success rate, this method is as robust as a terminal action: on one hand, the first point accounts for the fact that any agent fulfilling its scope has, at least to some degree, worked correctly; on the other hand, to avoid validating agents that accidentally enter a cave, we need to introduce a time threshold that clarifies an agent's intent.

Additionally, we highlight that examples featuring such terminal action would represent at most one action per episode. That is, terminal actions are a minimal fraction of the total number of actions in the dataset. Large models, such as VPT, fine-tuned on a relatively tiny amount of data (such as 10-100 trajectories) might fail to learn it. On the contrary, S-BC could make use of a single instance of terminal action, given the right circumstances. Therefore, keeping a terminal action would make the comparison between agents unfair. In introducing the five seconds threshold, our agent is the only one affected by our evaluation criterion.

To address the success rate problem, we make use of the cave classifier described in Appendix B2. In our evaluation, such classifier was not used as the only way of determining whether an episode was successful. Being aware of the potential catastrophic implications of a false positive (or false negative) in this context, we have manually evaluated each episode. To this end, the cave classifier proved to be a nice tool in easing the retrieval of potentially interesting segments of videos.
\section{Appendix - Ablation study}
S-BC heavily relies on the number of trajectories encoded in its search space: each point in the latent space corresponds to a potential solution to the current situation. Therefore, being able to control the shape and the density of such space is of the utmost importance. We tested the capabilities and limitations of our model over the number of encoded trajectories. In this experiment, we let each agent play $3$ runs of $30$ episodes for each task of the MineDojo suite and the FindCave task from BASALT. We assess each agent following the evaluation procedure detailed in Section 4. 

\renewcommand{\thefigure}{12}
\begin{figure*}[!t]
    \centering
    \includegraphics[width=\textwidth]{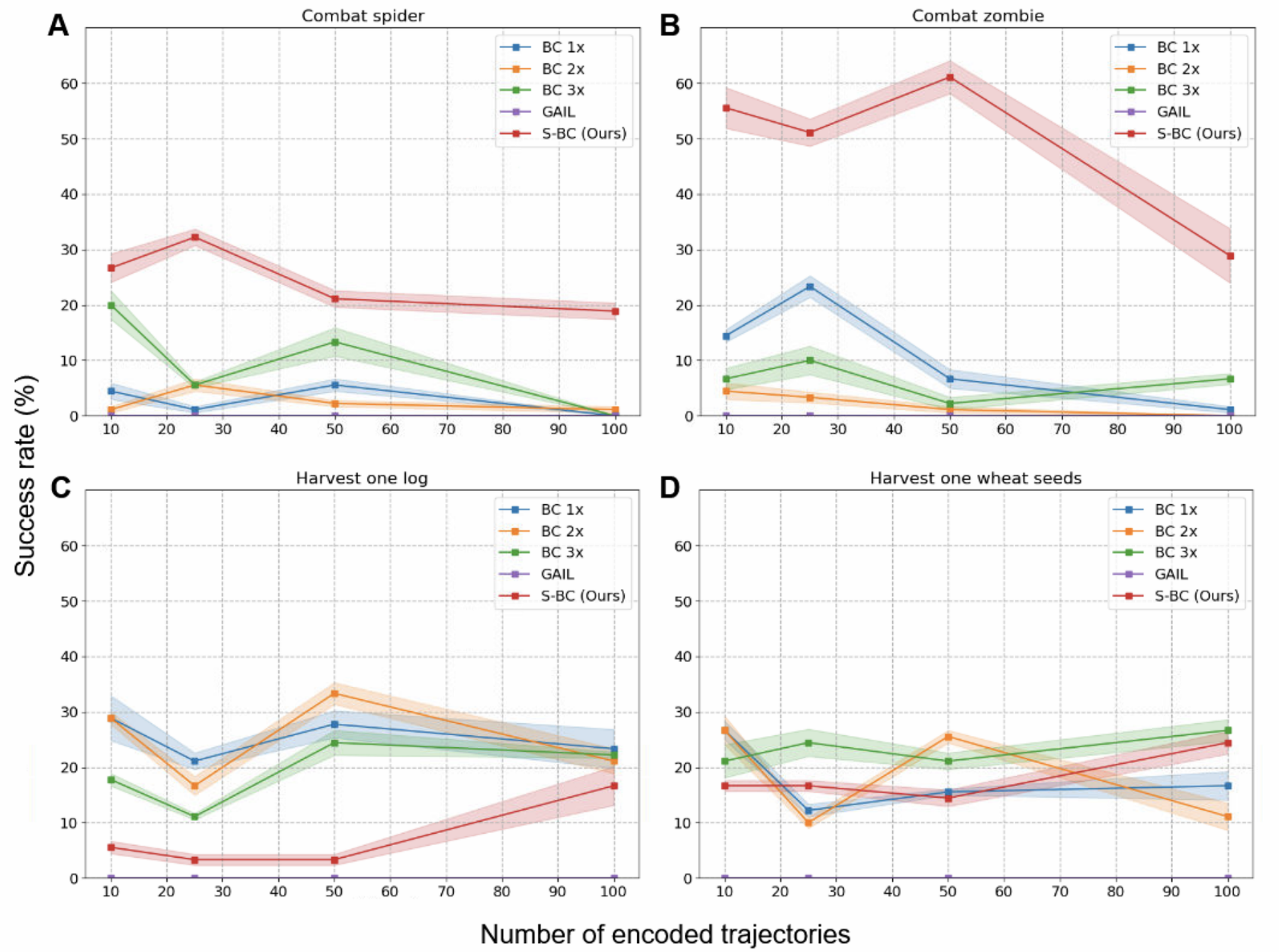}
    \caption{Results of the ablation study on selected MineDojo tasks. \textbf{(A) Combat spider:} an agent must fight and successfully kill a spider before the time limit; \textbf{(B) Combat zombie:} an agent must overcome a zombie unit; \textbf{(C) Harvest one log:} an agent must successfully gather one unit of wood from a nearby tree; \textbf{(D) Harvest one wheat seed:} an agent must harvest one unit of wheat seed from the ground.}
    \label{fig:ablation_dojo}
\end{figure*}

\renewcommand{\thefigure}{13}
\begin{figure*}[!t]
    \centering
    \includegraphics[width=\textwidth]{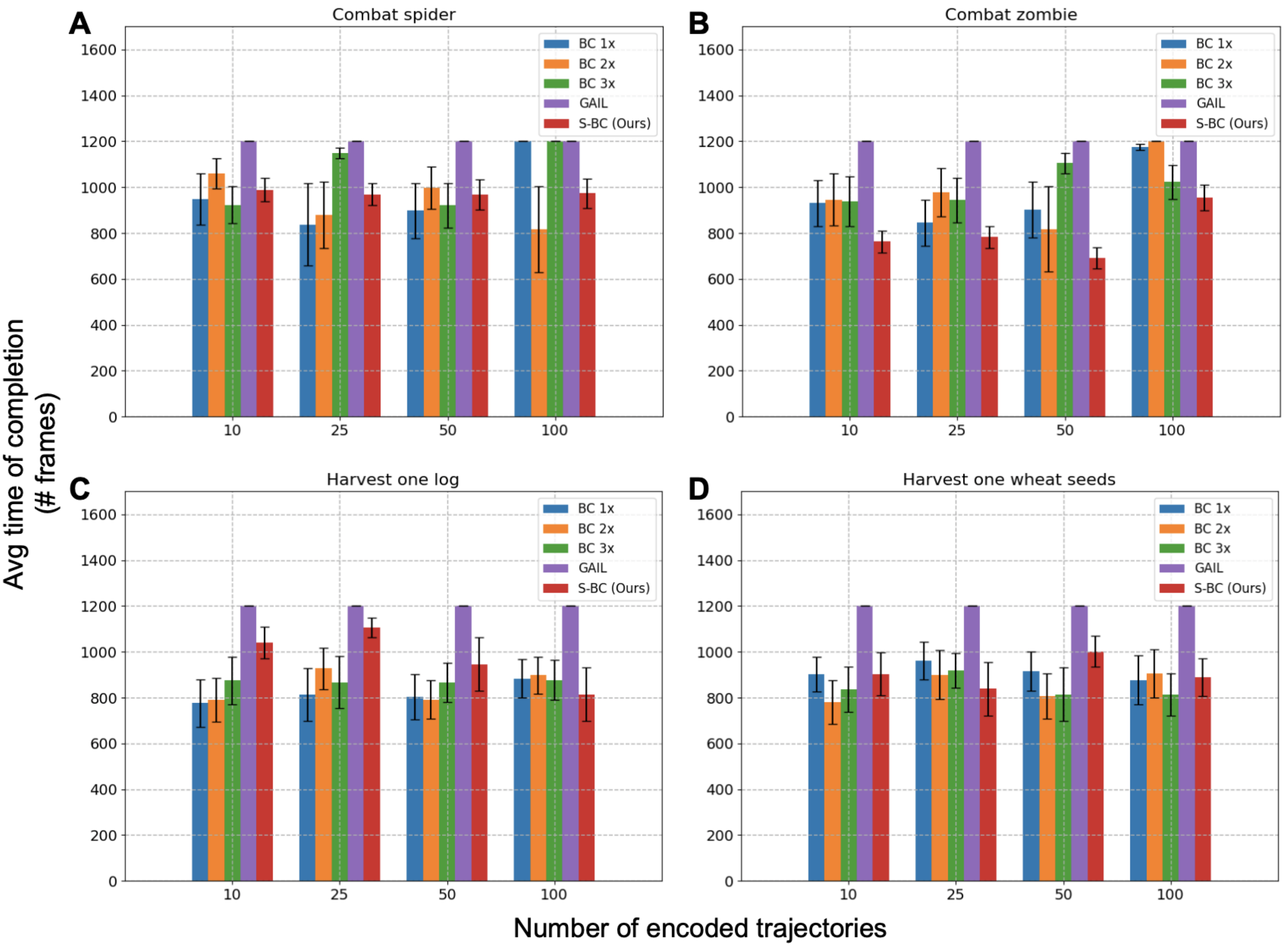}
    \caption{Average time of completion for ground truth tasks from the MineDojo suite \textbf{(A)} "Combat spider", \textbf{(B)} "Combat zombie", \textbf{(C)} "Harvest one log" and \textbf{(D)} "Harvest one wheat seed".}
    \label{fig:time_of_completion}
\end{figure*}

\renewcommand{\thetable}{D}
\begin{table*}
\centering
{\addtolength{\tabcolsep}{0pt}
\begin{tabular}{c c c c c c c}
  \textbf{\#} & \textbf{Task} & \textbf{BC 1x} & \textbf{BC 2x}  & \textbf{BC 3x} & \textbf{GAIL} & \textbf{S-BC (Ours)} \\[2pt] \hline \\[-7pt]

$10$ & Find Ocean & $20.46\pm1.16$ & $20.28\pm1.09$ & $19.74\pm1.00$ & $\textbf{20.60}\pm0.89$ & $20.31\pm1.12$\\[2pt] 
& Dig Hole                               & $21.15\pm0.91$                              & $20.96\pm0.84$     & $21.22\pm0.87$ & 
$\textbf{21.58}\pm1.10$ &  $20.81\pm0.90$   \\[2pt] \hline \\[-7pt]
$25$ & Find Ocean & $20.31\pm1.07$ & $20.26\pm1.05$ & $20.12\pm1.12$ & $\textbf{20.41}\pm0.90$ & $\textbf{20.41}\pm1.19$\\[2pt] 
& Dig Hole                               & $20.84\pm0.83$                              & $\textbf{20.95}\pm0.92$     & $20.91\pm0.79$ & 
$20.09\pm1.02$ &  $20.83\pm0.94$   \\[2pt] \hline \\[-7pt]
$50$ & Find Ocean & $\textbf{20.54}\pm1.21$ & $20.24\pm0.98$ & $20.05\pm1.10$ & $20.51\pm1.02$ & $20.33\pm1.20$\\[2pt] 
& Dig Hole                               & $20.73\pm0.84$                              & $20.83\pm0.82$     & $21.01\pm0.85$ & 
$\textbf{21.44}\pm0.80$ &  $20.96\pm0.86$   \\[2pt] \hline \\[-7pt]
$100$ & Find Ocean & $20.25\pm1.10$ & $20.19\pm1.04$ & $19.82\pm1.05$ & $20.30\pm0.74$ & $\textbf{20.47}\pm1.11$\\[2pt]
& Dig Hole                               & $20.80\pm0.88$                              & $20.94\pm0.85$     & $20.95\pm0.87$ & 
$20.20\pm0.24$ &  $\textbf{20.97}\pm0.92$                               \\[2pt] %
\end{tabular}
\caption{Results of the ablation study over available expert trajectories for the evaluated \textit{creative} tasks. Average MineCLIP score and its standard deviation shown per model and available expert trajectories. MineCLIP scores can be used to train agents in the MineDojo suite in an RL fashion.}
}
\label{tab:mineclipscores_ablation}
\end{table*}

\paragraph{MineDojo tasks} We trained each model described in main text using $10$, $25$, $50$ and $100$ trajectories. Results of the study are reported in Figure \ref{fig:ablation_dojo}, Figure \ref{fig:time_of_completion} and Table \ref{tab:mineclipscores_ablation}. 

In Figure \ref{fig:ablation_dojo} we show that S-BC reaches better or similar performance to the VPT-based models and GAIL in all the tasks, under almost all conditions. This finding support our claim of zero-shot capabilities through mimicking of experts' trajectories, with no requirement for training.  

Notably, Figure \ref{fig:ablation_dojo}A and B show that our agent is the only one capable of consistently solving combat-related tasks. In particular, we notice that S-BC achieves very high success rate ($50-60\%$) on \textit{Combat zombie} when using $10$, $25$ or $50$ trajectories, while performance degrade on the $100$ trajectories. We hypothesize that such behavior might occur when the density of information in latent space is too high with respect to the complexity of the task. Nonetheless, we plan to conduct further studies on this aspect. 

S-BC struggles in harvesting log (Figure \ref{fig:ablation_dojo}C) when relying on a low number of trajectories, while all the BC models reach around $30\%$ of success rate on the task. It is interesting to notice that harvesting wood naturally matches the aim of a VPT-based model. Therefore, we expected strong performances from these models on this task. Surprisingly, S-BC compensates a significant portion of this gap when using $100$ encoded trajectories. Given the dynamics of the task, we hypothesize that having more trajectories improves our agent's navigation skill to a point where it become competitive with VPT-based BC models. 

Navigation-related problems are partially solved when harvesting a wheat seed (Figure \ref{fig:ablation_dojo}D), as these item are quite common on the ground of certain biomes, and require therefore much weaker navigation skills. In this task, all the agents perform similarly and seem to be independent from the number of trajectories used.

It is interesting to notice how GAIL is completely unable to perform any of the tasks. We hypothesize that the specific nature of the tasks and the limited amount of data used was not representative enough to successfully train it, despite being the method that required the highest amount of time to be trained.

Evaluation on the creative tasks \textit{Find ocean} and \textit{Dig a hole} (Table \ref{tab:mineclipscores_ablation}) revealed that all the models achieve roughly the same performance. The loose specification of the goal for those tasks poses a hard problem in describing the absolute performance of an agent. Despite this, we highlight the highest score for each situation. S-BC achieves the best performance in \textit{Find ocean} with 25 trajectories, and in both \textit{Find ocean} and \textit{Dig a hole} with 100 encoded trajectories. Notably, GAIL yields similar performance and receives the highest rewards in four occasions, namely \textit{Find ocean} and \textit{Dig a hole} with 10 trajectories, \textit{Find ocean} with 25 trajectories and \textit{Dig a hole} with 50 trajectories. Perhaps surprisingly, BC 3x appears to be consistently worse than any other agent on the \textit{Find ocean} task.

\renewcommand{\thefigure}{14}
\begin{figure*}[!t]
    \centering
    \includegraphics[width=0.5\textwidth]{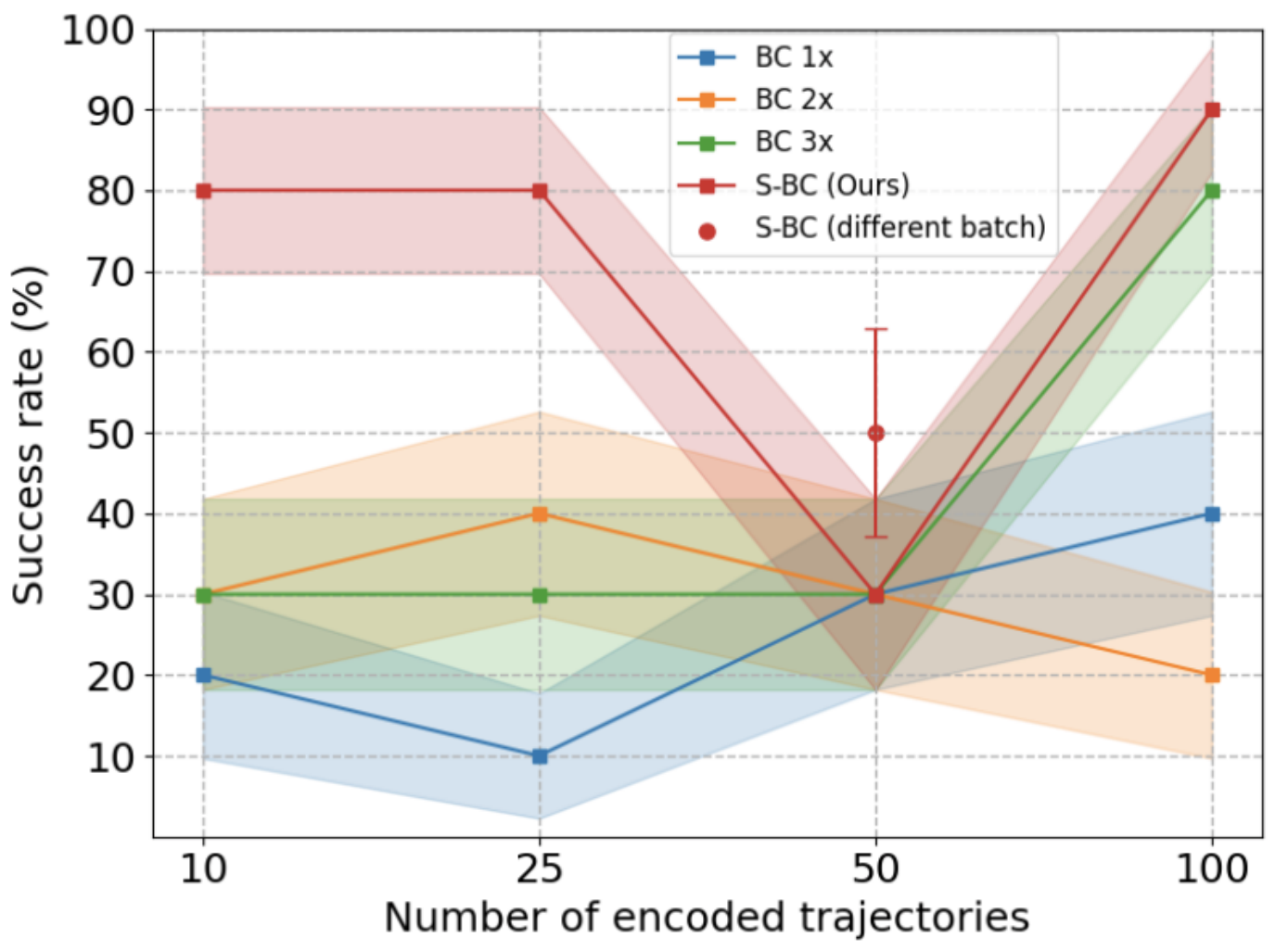}
    \caption{Ablation study over the number of trajectories in the FindCave scenario. We encoded $10$, $25$, $50$ and $100$ trajectories for S-BC, and used the same subset of trajectories to fine-tune several VPT foundation models. Point in red refers to a different batch of 50 trajectories used for S-BC.}
    \label{fig:success_findcave}
\end{figure*}

\paragraph{FindCave task}
Figure~\ref{fig:success_findcave} shows how the number of encoded trajectories affect our agent performance on the FindCave task. We have trained four versions of our model, using a subset of $10$, $25$, $50$ and $100$ trajectories from the original FindCave dataset. For comparison, we have fine-tuned a number of VPT-based foundation models using the same subset of trajectories. In all cases, our agent achieves baseline or better performance. More in detail, our agent dramatically improves performance with as low as $10$ and $25$ encoded trajectories, and reaches top performance with $100$ encoded trajectories, where it reaches $90\%$ of success rate. 

During our study, we noticed that the in the case of $50$ trajectories our agent's performance were degraded. We hypothesize that, since S-BC is heavily reliant on the quality of latents, such drop would correspond to a low quality batch of trajectories. Therefore, we re-trained our agent with a different batch or trajectories and stated an increase in performance of $20\%$. While this change did not result in similar performance to the other cases, it supports the fact that our agent's performance may significantly vary according to the quality of data used to generate the latent space.
\end{document}